\documentclass{ifacconf}

\counterwithin*{section}{part}
\usepackage{natbib}        
\usepackage[utf8]{inputenc} 
\usepackage[T1]{fontenc}    
\usepackage{url}            
\usepackage{amsfonts}       
\usepackage{amsmath}
\usepackage{nicefrac}       
\usepackage{microtype}      
\usepackage{cleveref}

\usepackage{wrapfig}
\usepackage{placeins}
\usepackage{subcaption}

\usepackage{booktabs}       
\usepackage{multicol}
\usepackage{multirow}
\usepackage{siunitx}
\usepackage{etoolbox}           
\renewcommand{\bfseries}{\fontseries{b}\selectfont} 
\robustify\bfseries             
\newrobustcmd{\B}{\bfseries} 

\usepackage{tikz}
\usetikzlibrary{shapes.geometric}
\usetikzlibrary{positioning}
\usetikzlibrary{calc}
\usetikzlibrary{fit}
\usetikzlibrary{backgrounds}
\usepackage{listofitems}
 
\usepackage{mymathsty}
\usepackage{dblfloatfix}
\usepackage{color}

\usepackage{listings}
\usepackage{times}

\begin{document}
	\newsavebox{\citethis}
	\begin{lrbox}{\citethis}
		\hspace{-5.5cm}
		\begin{minipage}{1.0\textwidth}
			\begin{lstlisting}[breaklines,basicstyle=\small\ttfamily]
			@inproceedings{CarlAndersson2021,
			author    = {Carl R. Andersson and Niklas Wahlstr\"{o}m and Thomas Bo Sch\"{o}n},
			title     = {Learning deep autoregressive models for hierarchical data},
			booktitle = {Proceedings of 19th IFAC Symposium on System Identification (SYSID)},
			address   = {Padova, Italy (online)},
			year      = {2021}
			}
			\end{lstlisting}
		\end{minipage}
	\end{lrbox}

\begin{frontmatter}

	\title{Learning deep autoregressive models for hierarchical data} 

	\author[First]{Carl R. Andersson} 
	\author[First]{Niklas Wahlström} 
	\author[First]{Thomas B. Schön}
	
	\address[First]{Department of information technology, Uppsala University, Sweden, (email: carl.andersson;niklas.wahlstrom;thomas.schon@it.uu.se)}
	
	{ 
		\vspace{2cm}
		\textbf{Please cite this version:}
		
		Carl R. Andersson, Niklas Wahlstr\"{o}m, Thomas B. Sch\"{o}n.   \textbf{Learning deep autoregressive models for hierarchical data}. In \textit{Proceedings of 19th IFAC Symposium on System Identification (SYSID)}, Padova, Italy (online),  2021.
		
		\usebox{\citethis}
		\vspace{2cm}
	}

	\begin{abstract}                
		We propose a model for hierarchical structured data as an extension to the stochastic temporal convolutional network. The proposed model combines an autoregressive model with a hierarchical variational autoencoder and downsampling to achieve superior computational complexity. We evaluate the proposed model on two different types of sequential data: speech and handwritten text. The results are promising with the proposed model achieving state-of-the-art performance.
	\end{abstract}

\end{frontmatter}

\begin{frontmatter}
	
	\title{Learning deep autoregressive models for hierarchical data} 
	
	\thanks[footnoteinfo]{
		This research was financially supported by the projects  \emph{Learning flexible models for nonlinear dynamics} (contract number: 2017-03807) by the Swedish Research Council, by \emph{AI4Research} at Uppsala University and by \emph{Kjell och Märta Beijer Foundation}.
	}
	
	\author[First]{Carl R. Andersson} 
	\author[First]{Niklas Wahlström} 
	\author[First]{Thomas B. Schön}
	
	\address[First]{Department of information technology, Uppsala University, Sweden, (email: carl.andersson;niklas.wahlstrom;thomas.schon@it.uu.se)}

	\begin{abstract}                
		We propose a model for hierarchical structured data as an extension to the stochastic temporal convolutional network. The proposed model combines an autoregressive model with a hierarchical variational autoencoder and downsampling to achieve superior computational complexity. We evaluate the proposed model on two different types of sequential data: speech and handwritten text. The results are promising with the proposed model achieving state-of-the-art performance.
	\end{abstract}
	
	
\end{frontmatter}

\section{Introduction}

System identification and sequence modeling with deep learning are two different research areas that essentially solve the same problem, to model a sequence, $y_{1:T}$, for predictive purposes. Even though the methods are very similar there are differences when it comes to what kind of data the method typically is applied to. Whereas system identification traditionally has focused on systems with relatively short memory and small datasets (e.g. systems on the nonlinear benchmark website \citep{NonLinear}), deep learning has focused on the opposite, i.e. systems with long memory and larger datasets (e.g. text modeling). This dichotomy is not a product of any fundamental difference in methodology, but rather an effect of the rapid advancement in computational power that has accompanied the advancements in deep learning.

\begin{figure}[t]
	\centering
	\begin{subfigure}[]{0.24\textwidth}
		\includegraphics[width=\linewidth, height=\linewidth]{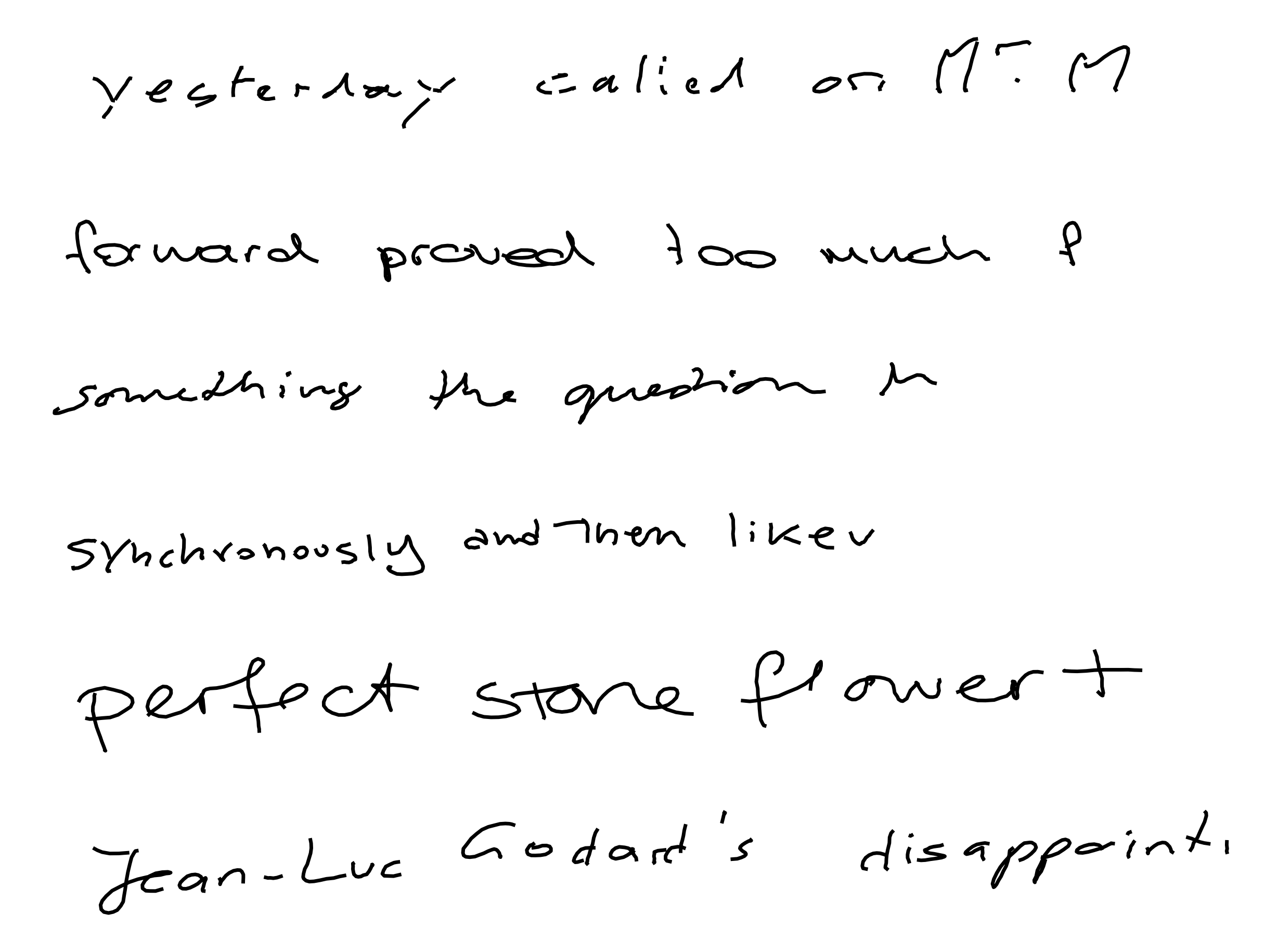}
		\caption{Dataset samples}
		\label{fig:data_samples}
	\end{subfigure}\hfill
	\begin{subfigure}[]{0.24\textwidth}
		\includegraphics[width=\linewidth, height=\linewidth]{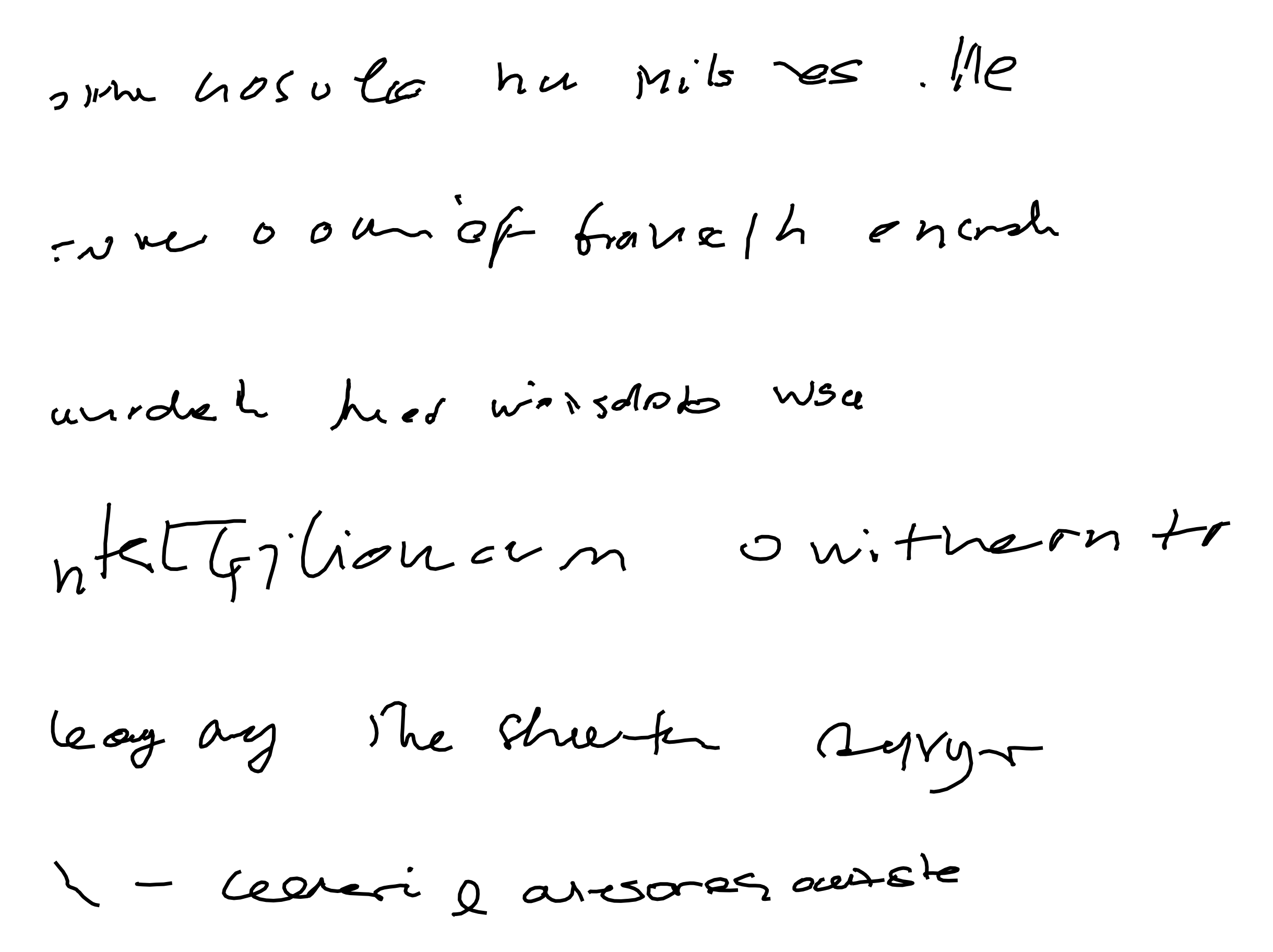}
		\caption{Generated samples}
	\end{subfigure}
	\caption{An illustration of samples from the dataset and generated samples from our model for the IAM-OnDB dataset \citep{Liwicki2005_iamondb}.}
	\label{fig:samples}
\end{figure}

Neural networks are by no means new to the system identification community, they have in fact a rather rich history \citep{Sjoberg1995}. What deep learning brought with it was instead a new paradigm of model design and implicit regularization. Whereas the system identification community typically view neural networks as black box function approximators, the deep learning community instead augment how these black box models are built up and adapt them to the data, e.g. long-short term memory (LSTM) models \citep{Hochreiter1997}, convolutional neural networks (CNN) \citep{Krizhevsky2012_AlexNet} and temporal convolutional networks (TCN). A very prominent example of the connection between model and data structure is indeed deep CNNs applied to images, where the model mirrors the hierarchical composition of natural images and  locality of low-level features \citep{LeCun2015} while high-level features are represented in a downsampled image. Such hierarchical composition can also be seen in many sequential datasets, e.g. language modeling, handwritten text modeling (see \Cref{fig:data_samples}), human motion tracking or other systems of (hierarchical) switching nature where the high-level features evolve at a slower pace than low-level features \citep{Chung2016_HMRNN, Koutnik2014}.

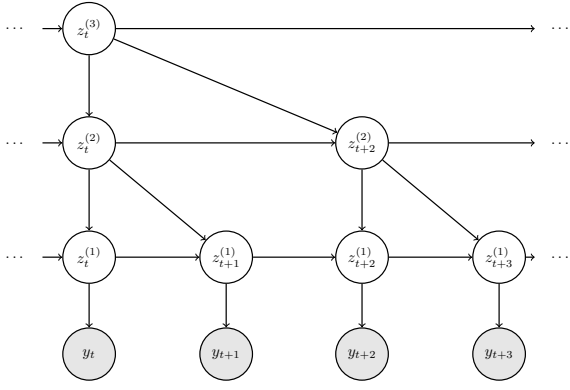
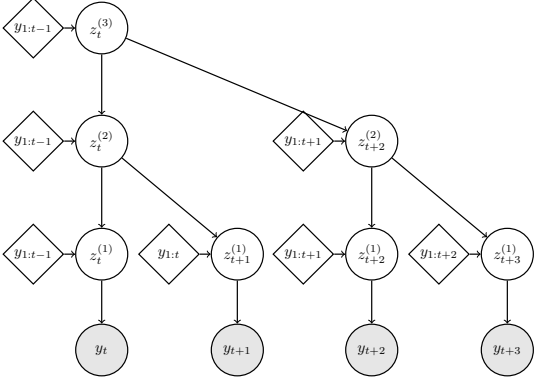
\begin{figure}[t]
	\centering
	\begin{subfigure}[t]{0.46\textwidth}
		\resizebox{!}{0.6\linewidth}{\begin{tikzpicture}
[align=center,node distance=1.5cm, minimum size=1.2cm]

\node[circle,thick,draw] (pz3) {$z^{(3)}_t$};

\node[circle,thick,draw,below = 1.4cm of pz3] (pz21) {$z^{(2)}_{t}$};
\node[circle,thick,draw,right = 5.0cm of pz21] (pz22) {$z^{(2)}_{t+2}$};

\node[circle,thick,draw,below = 1.4cm of pz21] (pz11) {$z^{(1)}_{t}$};
\node[circle,thick,draw,right = 1.9cm of pz11] (pz12) {$z^{(1)}_{t+1}$};

\node[circle,thick,draw,below = 1.4cm of pz22] (pz13) {$z^{(1)}_{t+2}$};
\node[circle,thick,draw,right = 1.9cm of pz13] (pz14) {$z^{(1)}_{t+3}$};

\node[circle,thick,draw,below=1.0cm of pz11,fill = black!10] (py1) {$y_{t}$};
\node[circle,thick,draw,below=1.0cm of pz12,fill = black!10] (py2) {$y_{t+1}$};
\node[circle,thick,draw,below=1.0cm of pz13,fill = black!10] (py3) {$y_{t+2}$};
\node[circle,thick,draw,below=1.0cm of pz14,fill = black!10] (py4) {$y_{t+3}$};

\node[left= 0.45cm of pz11] (pz1dpre) {$\dots$};
\node[right= 0.2cm of pz14] (pz1dpost) {$\dots$};
\node[]  at (pz3 -| pz1dpre)  (pz3dpre) {$\dots$};
\node[]  at (pz3 -| pz1dpost)  (pz3dpost) {$\dots$};
\node[]  at (pz21 -| pz1dpre)  (pz2dpre) {$\dots$};
\node[]  at (pz21 -| pz1dpost)  (pz2dpost) {$\dots$};

\draw[->, thick] (pz3dpre) edge (pz3);
\draw[->, thick] (pz3) edge (pz3dpost);

\draw[->, thick] (pz2dpre) edge (pz21);
\draw[->, thick] (pz22) edge (pz2dpost);

\draw[->, thick] (pz1dpre) edge (pz11);
\draw[->, thick] (pz14) edge (pz1dpost);

\draw[->,thick] (pz3) edge (pz21);
\draw[->,thick] (pz3) edge (pz22);

\draw[->,thick] (pz21) edge (pz11);
\draw[->,thick] (pz21) edge (pz12);

\draw[->,thick] (pz21) edge (pz22);

\draw[->,thick] (pz22) edge (pz13);
\draw[->,thick] (pz22) edge (pz14);

\draw[->,thick] (pz11) edge (py1);		
\draw[->,thick] (pz11) edge (pz12);

\draw[->,thick] (pz12) edge (py2);
\draw[->,thick] (pz12) edge (pz13);

\draw[->,thick] (pz13) edge (py3);
\draw[->,thick] (pz13) edge (pz14);

\draw[->,thick] (pz14) edge (py4);

\end{tikzpicture}}
		\caption{State space form}
		\label{fig:deep_seq}
	\end{subfigure}
	\hspace{0.02\textwidth}
	\begin{subfigure}[t]{0.46\textwidth}
		\resizebox{!}{0.6\linewidth}{\begin{tikzpicture}
[align=center,node distance=1.5cm, minimum size=1.2cm]

\node[circle,thick,draw] (pz3) {$z^{(3)}_{t}$};

\node[circle,thick,draw,below = 1.4cm of pz3] (pz21) {$z^{(2)}_{t}$};
\node[circle,thick,draw,right = 5.0cm of pz21] (pz22) {$z^{(2)}_{t+2}$};

\node[circle,thick,draw,below = 1.4cm of pz21] (pz11) {$z^{(1)}_{t}$};
\node[circle,thick,draw,right = 1.9cm of pz11] (pz12) {$z^{(1)}_{t+1}$};

\node[circle,thick,draw,below = 1.4cm of pz22] (pz13) {$z^{(1)}_{t+2}$};
\node[circle,thick,draw,right = 1.9cm of pz13] (pz14) {$z^{(1)}_{t+3}$};

\node[circle,thick,draw,below=1.0cm of pz11,fill = black!10] (py1) {$y_{t}$};

\node[circle,thick,draw,below=1.0cm of pz12,fill = black!10] (py2) {$y_{t+1}$};

\node[circle,thick,draw,below=1.0cm of pz13,fill = black!10] (py3) {$y_{t+2}$};

\node[circle,thick,draw,below=1.0cm of pz14,fill = black!10] (py4) {$y_{t+3}$};

\node[diamond,thick,draw,left = 0.25cm of pz3, inner sep=1pt, minimum size=1.4cm] (d3) {$y_{1:t-1}$};

\node[diamond,thick,draw,left = 0.25cm of pz21, inner sep=1pt, minimum size=1.4cm] (d21) {$y_{1:t-1}$};
\node[diamond,thick,draw,left = 0.25cm of pz22, inner sep=1pt, minimum size=1.4cm] (d22) {$y_{1:t+1}$};

\node[diamond,thick,draw,left = 0.25cm of pz11, inner sep=1pt, minimum size=1.4cm] (d11) {$y_{1:t-1}$};
\node[diamond,thick,draw,left = 0.25cm of pz12, inner sep=1pt, minimum size=1.4cm] (d12) {$y_{1:t}$};
\node[diamond,thick,draw,left = 0.25cm of pz13, inner sep=1pt, minimum size=1.4cm] (d13) {$y_{1:t+1}$};
\node[diamond,thick,draw,left = 0.25cm of pz14, inner sep=1pt, minimum size=1.4cm] (d14) {$y_{1:t+2}$};

\node[right= 0.2cm of pz14] (pz1dpre) { };

\draw[->,thick] (d3) edge (pz3);

\draw[->,thick] (pz3) edge (pz21);
\draw[->,thick] (pz3) edge (pz22);

\draw[->,thick] (d21) edge (pz21);

\draw[->,thick] (pz21) edge (pz11);
\draw[->,thick] (pz21) edge (pz12);

\draw[->,thick] (d22) edge (pz22);

\draw[->,thick] (pz22) edge (pz13);
\draw[->,thick] (pz22) edge (pz14);

\draw[->,thick] (d11) edge (pz11);

\draw[->,thick] (pz11) edge (py1);		

\draw[->,thick] (d12) edge (pz12);

\draw[->,thick] (pz12) edge (py2);

\draw[->,thick] (d13) edge (pz13);

\draw[->,thick] (pz13) edge (py3);	

\draw[->,thick] (d14) edge (pz14);

\draw[->,thick] (pz14) edge (py4);

\end{tikzpicture}} \hfill
		\caption{Autoregressive form}
		\label{fig:deep_seq_ar_v1}
	\end{subfigure}
	\caption{Two versions of a hierarchical generative model for sequential data, $y$, on state space form (top) and autoregressive form (bottom), with latent variables, $z$. $z^{(2)}$ updates only every other timestep and $z^{(3)}$ updates only every forth. }
\end{figure}

In the light of this we propose a sequential hierarchical latent variable model, represented in \Cref{fig:deep_seq}, that focus on modeling multiscale hierarchically structured sequential data. This can seen as an extension the Stochastic TCN \citep{Aksan2019} which also uses a sequential hierarchical latent variable model although without the multiscale property. The latent variables $z= \{z^{(l)}_{1:T}\}$ are structured sequentially with hierarchical layers, where layer $l$ is downsampled with a factor $S^{(l)}$. To facilitate training we propose to relax the state space formulation similarily to \citet{Aksan2019}, see \Cref{fig:deep_seq_ar_v1}. With this we get a deep sequential model that exploits the hierarchical nature of the data similarly to how a deep convolutional network exploits the nature of images. We realize the model using a combination of a Wavenet model \citep{VanDenOord2016} for the sequential component with a hierarchical variational autoencoder \citep{Rezende2014,Kingma2014} for the latent variables.  We employ the proposed model on two different datasets and show that by including the prior information of the hierarchical nature of the data we can improve the model in terms of both parameter effectiveness and computational complexity.

\section{Background}

In this section we will introduce and give some background on the components needed to realize the proposed model.

\subsection{Temporal convolutional network}
\vspace{-0.5em}

The temporal Convolutional Network (TCN) is a type of non-linear autoregressive (NARX) model that makes use  of a deep autoregressive neural network model. \cite{Bai2018} observe that these autoregressive models are easier to train, less sensitive to hyperparameter selection and in many circumstances achieves superior results, compared to deterministic state space models, i.e. LSTMs and RNNs. This can partly be explained by the efficient implementation, both computationally and parameterwise, of the  autoregressive neural networks here realized with convolutions.

The depth of the TCN is achieved by stacking multiple affine transformations intertwined with activation functions. Each affine transformation/activation function pair can be expressed with
\begin{align}
	h_t = \phi(x_{t-k:t-1}),
\end{align}
where $\phi$ combines an affine transformation and a nonlinear activation function of the regression vector, $x_{t-k:t-1}$. The notation $x_{i:j}$ denotes the sequence of $x$ from $i$ to $j$. An alternative formulation of this is to view the entire $h_{1:T}$ as a convolution between $\phi$ and the input $x_{1:T}$. The width of this convolution filter, $k$, corresponds to the so called the \emph{receptive field} of the convolution. We use the notation 
\begin{align}
	h_{1:T} = f(\text{Conv}(x_{1:T})),
\end{align}
where Conv stands for the affine transformation and $f$ is an activation function. 

The dilated convolution \citep{VanDenOord2016} is an option to the convolution that sets some of the parameters in the affine transformation to zero. With a dilation rate $\mu$ only every $\mu$:th element in the affine transformation is nonzero. By convention $k$ denotes the number of effective parameters in this transformation. Thus, this increase the receptive field of the filter to $k \mu$. By stacking several dilated convolutions with exponentially increasing dilation rate it is possible to produce a model with exponentially long memory (exponentially large respective field) with a linear increase in the number of parameters \citep{Yu2016}. The parameters of the affine transformation are the parameters of the convolutional layer, while both $k$ and $\mu$ are hyperparameters. 

\subsection{Wavenet}
\vspace{-0.5em}

One of the most prominent TCNs is Wavenet \citep{VanDenOord2016}. Wavenet is built up of stacked so-called Wavenet blocks, where each block consists of a residual connection and an identity connection (inspired by  ResNet \citep{He2016}).  Each residual connection in turn consists of a sequence of: a dilated convolution, a gate-like activation function, and a final $1{\times}1$ convolution. An $1{\times}1$ convolution is a type of convolution that only operates locally (the filter has size 1). The Wavenet block can be summarized as
\begin{align}
\text{WavenetBlock}(x_{1:T}) &= x_{1:T} +  \nonumber \\ 
\text{Conv}_{1{\times}1} (
\tanh(&\text{Conv}(x_{1:T})) \odot 
\sigma(\text{Conv}(x_{1:T}))),
\end{align} 
where $\odot$ denotes element-wise multiplication and $\sigma$ denotes the sigmoid activation function.

When merging two data sequences with the Wavenet model \cite{VanDenOord2016} proposed to consider one of the sequences as the main sequence ($x_{1:T}$) and the other as a conditional sequence ($c_{1:T}$). The two sequences are concatenated (in feature space) forming a new stream and this is used as input to the residual network. The identity connection only passes $x_{1:T}$ forward. Thus, a conditional Wavenet block can be written as,
\begin{align}
	&\text{CondWavenetBlock}(x_{1:T},c_{1:T}) = x_{1:T} +  \nonumber \\ &\text{Conv}_{1{\times}1}(\tanh(\text{Conv}([x_{1:T},c_{1:T}])) \odot \sigma(\text{Conv}([x_{1:T},c_{1:T}]))) ,
\end{align} 
where $[\cdot]$ denotes concatenation in the feature dimension. When stacking several such blocks together, all blocks share the same conditioning sequence. \citet{Lai2018} with Stochastic Wavenet and later \citet{Aksan2019} extended these models to also include latent variables similar to the proposed model. 

\subsection{Variational autoencoder}
\vspace{-0.5em}

\label{sec:VAE}
A common approach for latent variables in deep learning is to use variational autoencoders (VAE) \citep{Rezende2014,Kingma2014}. This section aims to introduce the notation we use in this paper and for a more in depth and pedagogical description of the VAE see \citet{Kingma2019}. The idea of a VAE is to model the distribution of $y$ with a generative model as
\begin{equation}
\label{eq:vae_gen}
p(y) = \expect \big  [ p(y \given z) p(z) \big ],
\end{equation}
with 
\begin{align*}
	p(y \given z) &= \mathcal{N}(y \given \mu_y(z), \Sigma_y(z)), \\ 
	p(z) &= \mathcal{N}(z\given 0, I),
\end{align*}
where $\mu_y(z)$ and $\Sigma_y(z)$ are modeled as neural networks. 

In the VAE setting the expectation in \eqref{eq:vae_gen} is estimated with Monte Carlo samples and (amortized) variational inference. The variational inference is done by introducing a parameterized approximate posterior distribution,~$q$, 
\begin{align*}
q(z) = \mathcal{N}(z \given \mu_q(y), \Sigma_q(y)),
\end{align*}
where $\mu_q(y)$ and $\Sigma_q(y)$ are also modeled as neural networks. It is also possible to condition the VAE on some arbitrary data, $c$, in which case the prior is similarly expressed with neural networks,
\begin{align}
p(z \given c) &= \mathcal{N}(z \given \mu_p(c), \Sigma_p(c)).
\end{align}
The approximate posterior $q(z)$ and the generative distribution will then also depend on $c$. 

Using this approximate posterior and Jensen's inequality, we can bound the log-likelihood from below. This gives us the evidence lower bound (ELBO)
\begin{equation}
\mathcal{L}_\text{ELBO}(y;\theta) =  \mathbb{E}_{q} \big [ \log p(y \given z) \big ] -  D_\text{KL} \left(q(z) || p(z) \right ) ,
\label{eq:elbo}
\end{equation}
where $D_\text{KL}$ denotes the Kullback-Lieber divergence and $\theta$ are the parameters of both the approximative posterior and the generative model. The expectation here is evaluated with respect to $q$ and it is most commonly approximated with a single Monte Carlo sample. The ELBO objective is used to update both the parameters of the generative model and the approximate posterior. 

The variational RNN (VRNN) \citep{Chung2015}, stochastic RNN (SRNN) \citep{Fraccaro2016} and STORN \citep{Bayer2014} are all examples of VAEs that have been adapted to sequential data. The core of these models is an RNN which is extended with a latent variable. These models require a backward flow for inference which is implemented by an RNN running backwards in time.

\subsection{Hierarchical variational autoencoders}
\vspace{-0.5em}

\begin{figure}[t]
	\centering
	\begin{subfigure}[t]{0.46\textwidth}
		\centering
		\hspace{0.4cm}\scalebox{0.7}{\begin{tikzpicture}
	[align=center,node distance=1.75cm,  mycircle/.style={minimum size=0.8cm, circle}, mydiamond/.style={diamond, minimum size=1.1cm} ]

	\coordinate (last) at (0,0);
	\def\basename{ARL}
	
	\def\layerlist{1/1,2/2,3/3}
	
	\foreach \layer/\layername in \layerlist{
	
		\def\name{\basename\layer};
		
		\coordinate (in\name) at (last);
		\node[mydiamond, draw=red,thick, above=2.7cm of in\name, inner sep=1pt] (d\name) {$d^{(\layername)}$};
		\ifthenelse{\equal{\name}{ARL1}}{
			\draw[->,thick,red] (in\name) to node[midway,left, align=left] () {} (d\name);
		}{
			\draw[->,thick,red] (in\name) to node[midway,left, align=left] () {} (d\name);
		}

		\node[mydiamond,draw,thick, right=2cm of d\name, inner sep=1pt] (h\name) {$h^{(\layername)}$};

		\coordinate (hdhalfway\name) at ($(d\name)!0.5!(h\name)$);
		\coordinate (hd3halfway\name) at ($(d\name)!1.5!(h\name)$);
		
		\node[mycircle, draw,thick, below=0.6cm of hdhalfway\name, draw=red, inner sep=1pt] (q\name) {$z^{(\layername)}$};	
		\node[mycircle, draw,thick, below=0.6cm of hd3halfway\name, draw=blue, inner sep=1pt] (p\name) {$z^{(\layername)}$};
		\draw[dotted,thick,<-] (p\name) edge node[below,near end] (KL\name) {KL} (q\name);
		
		\draw[->,thick, red] (d\name) to (q\name);
		\draw[->,thick, red] (h\name) to (q\name);
		\draw[->,thick, blue] (h\name) to (p\name);
		
		\coordinate[below=1.4cm of h\name] (hzmerge\name);

		\coordinate (g\name) at (in\name -| hzmerge\name) ;
		\ifthenelse{\equal{\name}{ARL1}}{
			\draw[thick, blue] (hzmerge\name) to node[midway,right,align=left] () {} (g\name);
			\draw[thick, blue] (p\name) edge (hzmerge\name);
			\draw[thick,dashed, blue]  (h\name) edge (hzmerge\name);
		}{
			\draw[thick] (hzmerge\name) to node[midway,right,align=left] () {} (g\name);
			\draw[thick] (p\name) edge (hzmerge\name);
			\draw[thick,dashed]  (h\name) edge (hzmerge\name);
		}
		
		\coordinate[alias=last, above=0.1cm of d\name] (out\name);
	}


	\begin{scope}[on background layer]
		\draw[thick,red] (dARL1) to (inARL2);
		\draw[thick,red] (dARL2) to (inARL3);
		\draw[->, thick] (gARL3) to (hARL2);
		\draw[->, thick] (gARL2) to (hARL1);
	\end{scope}

	\node[circle,thick,draw,below=0.1cm of inARL1,fill = black!10] (qy) {$y$};	
	\node[circle,thick,draw,below=0.1cm of gARL1.center,fill = black!10] (py) {$y$};	

	\draw[thick,red] (qy) edge (inARL1);
	\draw[->,thick, blue] (gARL1) to (py);
	
;	\end{tikzpicture}}
		\caption{Hierarchical VAE. The vertical dashed line was not present in the original ladder VAE \citep{Sonderby2016} but it was used in the ResNet VAE \citep{Kingma2016}. }
		\label{fig:ladder_network}
	\end{subfigure}

	\begin{subfigure}[t]{0.46\textwidth}
		\centering
		\scalebox{0.7}{\begin{tikzpicture}
	[align=center,node distance=1.75cm, mycircle/.style={minimum size=0.8cm, circle}, mydiamond/.style={diamond, minimum size=1.1cm} ]

	\coordinate (last) at (0,0);
	\def\basename{ARL}
	
	\def\layerlist{1/1,2/2,3/3}
	
	\foreach \layer/\layername in \layerlist{
	
		\def\name{\basename\layer};
		
		\coordinate (in\name) at (last);
		\node[mydiamond, draw,thick, above=2.7cm of in\name, inner sep=1pt] (d\name) {$d^{(\layername)}$};
		\ifthenelse{\equal{\name}{ARL1}}{
			\draw[->,thick] (in\name) to node[midway,left, align=left] () {Wavenet} (d\name);
		}{
			\draw[->,thick] (in\name) to node[midway,left, align=left] () {Wavenet\\Downsample} (d\name);
		}

		\node[mydiamond,draw,thick, right=2cm of d\name, inner sep=1pt] (h\name) {$h^{(\layername)}$};
		\draw[->,thick,dashed] (d\name) to node[midway,above=-0.1cm] () {Delay} (h\name);

		\coordinate (hdhalfway\name) at ($(d\name)!0.5!(h\name)$);
		\coordinate (hd3halfway\name) at ($(d\name)!1.5!(h\name)$);
		
		\node[mycircle, draw,thick, below=0.6cm of hdhalfway\name, draw=red, inner sep=1pt] (q\name) {$z^{(\layername)}$};	
		\node[mycircle, draw,thick, below=0.6cm of hd3halfway\name, draw=blue, inner sep=1pt] (p\name) {$z^{(\layername)}$};
		\draw[dotted,thick,<-] (p\name) edge node[below,near end] (KL\name) {KL} (q\name);
		
		\draw[->,thick, red] (d\name) to (q\name);
		\draw[->,thick, red] (h\name) to (q\name);
		\draw[->,thick, blue] (h\name) to (p\name);
		
		\coordinate[below=1.4cm of h\name] (hzmerge\name);

		\coordinate (g\name) at (in\name -| hzmerge\name) ;
		\ifthenelse{\equal{\name}{ARL1}}{
			\draw[thick, draw=blue] (hzmerge\name) to node[midway,right,align=left] () {1x1 CondWavenet} (g\name);
			\draw[thick, blue] (p\name) edge (hzmerge\name) (h\name) edge (hzmerge\name);
		}{
			\draw[thick] (hzmerge\name) to node[midway,right,align=left] () {1x1 CondWavenet\\Upsample\\1x1 Wavenet} (g\name);
			\draw[thick] (p\name) edge (hzmerge\name) (h\name) edge (hzmerge\name);
		}
		
		\coordinate[alias=last, above=0.1cm of d\name] (out\name);
	}


	\begin{scope}[on background layer]
		\draw[thick] (dARL1) to (inARL2);
		\draw[thick] (dARL2) to (inARL3);
		\draw[->, thick] (gARL3) to (hARL2);
		\draw[->, thick] (gARL2) to (hARL1);
	\end{scope}

	\node[circle,thick,draw,below=0.1cm of inARL1,fill = black!10] (qy) {$y$};	
	\node[circle,thick,draw,below=0.1cm of gARL1.center,fill = black!10] (py) {$y$};	

	\draw[thick] (qy) edge (inARL1);
	\draw[->,thick, blue] (gARL1) to (py);
	
;	\end{tikzpicture}}
		\caption{The proposed model, the horizontal dashed line enables the useful autoregressive relationship. }
		\label{fig:our_ladder_network}
	\end{subfigure}
	\caption{\textbf{\textcolor{red}{Approximative posterior model}}, \textbf{\textcolor{blue}{generative model}} and \textbf{common features} for hierarchical VAEs (\subref{fig:ladder_network}) and our model (\subref{fig:our_ladder_network}). The dotted line encodes the KL divergence between the prior distribution and the approximate posterior. Note that during likelihood evaluation $z^{(l)}$ is sampled from the posterior.}
\end{figure}
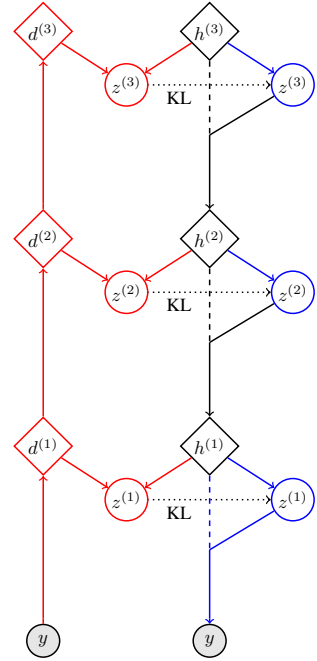
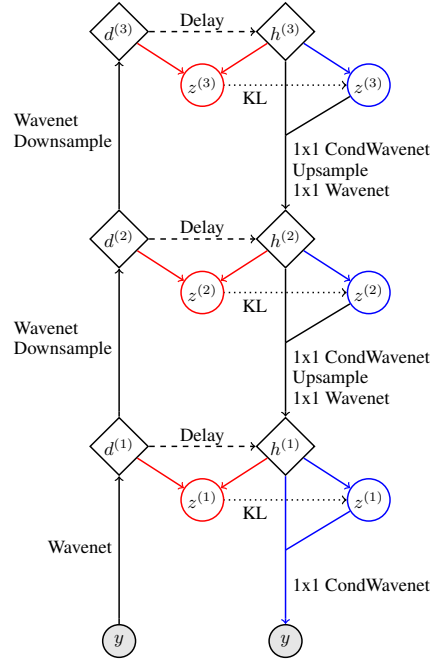

A hierarchical variational autoencoder extends the VAE by introducing hierarchically stacked latent variables, $\{z^{(l)} \}_{l=1}^L$, in $L$ layers. The use of stacked latent variables was proposed along with the original variational autoencoder, though with small improvements compared to the single layer VAE. The hierarchical models were later improved by \citet{Sonderby2016} with the so called ladder VAE, where the approximate posterior is linked to the generative distribution (see \Cref{fig:ladder_network}). This idea has since then proved to be fertile ground for numerous other architectures, e.g. ResNet VAE \citep{Kingma2016}, BIVA \citep{Maaloe2019} and NVAE \citep{vahdat2020}. 

The approximate posterior's dependence on $y$ is implemented by extracting a bottom-up hierarchy of features, $d^{(l)}$, where each successive layer of features depends on the previous layer, $d^{(l-1)}$. The link between the generative and the approximate posterior distribution has proven to be essential for efficient training of these models. Although, the implementation details for this link varies it can represented as \Cref{fig:ladder_network} in that the posterior also depends on the extracted features $h^{(l)}$ from the generative model. The parameters of the approximate posterior are optimized jointly with the parameters of the generative model.

\vspace{-0.5em}
\section{Model}
\vspace{-0.5em}

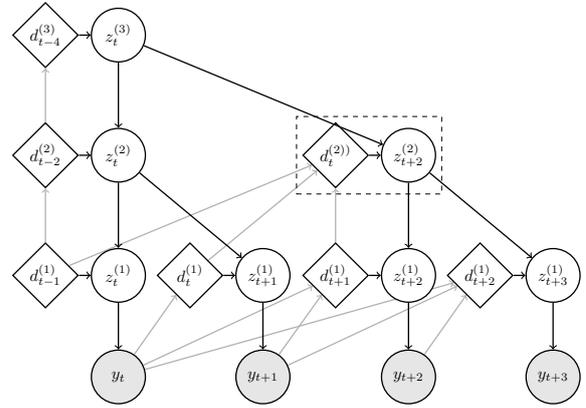
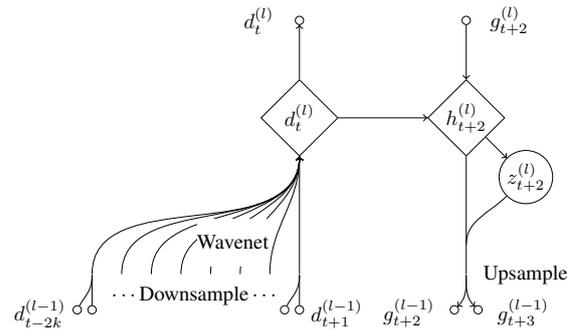
\begin{figure}[ht]
	\centering
	
	\begin{subfigure}[t]{0.46\textwidth}
		\resizebox{0.9\textwidth}{!}{\begin{tikzpicture}
[align=center,node distance=1.5cm, minimum size=1.15cm, every node/.style={fill=white}]

\node[circle,thick,draw] (pz3) {$z^{(3)}_{t}$};

\node[circle,thick,draw,below = 1.4cm of pz3] (pz21) {$z^{(2)}_{t}$};
\node[circle,thick,draw,right = 5.0cm of pz21] (pz22) {$z^{(2)}_{t+2}$};

\node[circle,thick,draw,below = 1.4cm of pz21] (pz11) {$z^{(1)}_{t}$};
\node[circle,thick,draw,right = 1.9cm of pz11] (pz12) {$z^{(1)}_{t+1}$};

\node[circle,thick,draw,below = 1.4cm of pz22] (pz13) {$z^{(1)}_{t+2}$};
\node[circle,thick,draw,right = 1.9cm of pz13] (pz14) {$z^{(1)}_{t+3}$};

\node[circle,thick,draw,below=1.0cm of pz11,fill = black!10] (py1) {$y_{t}$};
\node[circle,thick,draw,below=1.0cm of pz12,fill = black!10] (py2) {$y_{t+1}$};
\node[circle,thick,draw,below=1.0cm of pz13,fill = black!10] (py3) {$y_{t+2}$};
\node[circle,thick,draw,below=1.0cm of pz14,fill = black!10] (py4) {$y_{t+3}$};

\node[diamond,thick,draw,left = 0.25cm of pz3, inner sep=1pt, minimum size=1.4cm] (d3) {$d^{(3)}_{t-4}$};

\node[diamond,thick,draw,left = 0.25cm of pz21, inner sep=1pt, minimum size=1.4cm] (d21) {$d^{(2)}_{t-2}$};
\node[diamond,thick,draw,left = 0.25cm of pz22, inner sep=1pt, minimum size=1.4cm] (d22) {$d^{(2))}_{t}$};

\node[diamond,thick,draw,left = 0.25cm of pz11, inner sep=1pt, minimum size=1.4cm] (d11) {$d^{(1)}_{t-1}$};
\node[diamond,thick,draw,left = 0.25cm of pz12, inner sep=1pt, minimum size=1.4cm] (d12) {$d^{(1)}_{t}$};
\node[diamond,thick,draw,left = 0.25cm of pz13, inner sep=1pt, minimum size=1.4cm] (d13) {$d^{(1)}_{t+1}$};
\node[diamond,thick,draw,left = 0.25cm of pz14, inner sep=1pt, minimum size=1.4cm] (d14) {$d^{(1)}_{t+2}$};

\node[draw, dashed, fit=(d22)(pz22), fill=none] () {};

\draw[->,thick] (d3) edge (pz3);

\draw[->,thick] (pz3) edge (pz21);
\draw[->,thick] (pz3) edge (pz22);

\draw[->,thick] (d21) edge (pz21);

\draw[->,thick] (pz21) edge (pz11);
\draw[->,thick] (pz21) edge (pz12);

\draw[->,thick] (d22) edge (pz22);

\draw[->,thick] (pz22) edge (pz13);
\draw[->,thick] (pz22) edge (pz14);

\draw[->,thick] (d11) edge (pz11);

\draw[->,thick] (pz11) edge (py1);		

\draw[->,thick] (d12) edge (pz12);

\draw[->,thick] (pz12) edge (py2);

\draw[->,thick] (d13) edge (pz13);

\draw[->,thick] (pz13) edge (py3);	

\draw[->,thick] (d14) edge (pz14);

\draw[->,thick] (pz14) edge (py4);

\begin{scope}[on background layer, every path/.style={thick}]
	\draw[->, color=black!30] (py1) edge (d12);
	\draw[->, color=black!30] (py2) edge (d13);
	\draw[->, color=black!30] (py3) edge (d14);
	
	\draw[->, color=black!30] (py1) edge (d13);
	\draw[->, color=black!30] (py1) edge (d14);
	
	\draw[->, color=black!30] (py2) edge (d14);

	\draw[->, color=black!30] (d11) edge (d21);
	
	\draw[->, color=black!30] (d11) edge (d22);
	\draw[->, color=black!30] (d12) edge (d22);
	\draw[->, color=black!30] (d13) edge (d22);
	
	\draw[->, color=black!30] (d21) edge (d3);

\end{scope}

\end{tikzpicture}}
		\caption{The hierarchical structure of the generative model for a sample with four timesteps. This structure repeats to the right for longer sequences. The gray arrows corresponds to the bottom-up paths. Note that the $d$s are delayed with $S^{(l)}$. }
		\label{fig:gen_struct}
	\end{subfigure}
	\hspace{0.05\textwidth}
	\begin{subfigure}[t]{0.46\textwidth}
		\resizebox{0.9\textwidth}{!}{\begin{tikzpicture}
[align=center, minimum size=0.8cm]

\begin{scope}[yscale=0.6, xscale=0.25, local bounding box=downsample]
\foreach \i in {0,1,2,3,4,5,6,7}{
	
	\ifthenelse{\i=7 \OR \i =0}{
		\node[circle, draw, minimum size=0pt, inner sep=0.0pt] (downsample\i) at (2*\i+1,1) {};
		\node[circle, draw, minimum size=4pt, inner sep=0.1pt] (downsample\i1) at (2*\i,0) {};
		\node[circle, draw, minimum size=4pt, inner sep=0.1pt] (downsample\i2) at (2*\i+1,0) {};
		\draw[] (downsample\i1) edge[in=-90] (downsample\i);
		\draw[] (downsample\i2) edge (downsample\i);
		}{
		\node[circle, minimum size=0pt, inner sep=0pt] (downsample\i) at (2*\i+1,1) {};
		\node[circle, minimum size=4pt, inner sep=0.1pt] (downsample\i1) at (2*\i,0) {};
		\node[circle, minimum size=4pt, inner sep=0.1pt] (downsample\i2) at (2*\i+1,0) {};
	}
}
\node[minimum size=0.1cm] () at (8,0.4) {$\cdots \text{Downsample} \cdots$};
\end{scope}

\node[left=0cm of downsample.south west] () {$d_{t-2k}^{(l-1)}$};

\node[right=0cm of downsample.south east] () {$d_{t+1}^{(l-1)}$};

\node[diamond,draw,minimum size=1.3cm, inner sep=1pt, above=2cm of downsample7] (d) {$d^{(l)}_t$};
\foreach \i in {0, ..., 7}{
	\draw[->] (downsample\i) edge[out=90, in=-90] (d.south);
}

\node[above right=0.2cm of downsample.north, fill=white, inner sep=0pt] () {Wavenet};

\node[circle, draw, minimum size=4pt, inner sep=0.1pt] (dout) at ($ (d.north) + (0,1) $) {};

\draw[->] (d.north) -> (dout);

\node[diamond, draw, right=1.5cm of d, minimum size=1.3cm, inner sep=0pt] (pd) {$h^{(l)}_{t+2}$};
\coordinate (ppd) at ($(pd.south) + (0,-1.5cm)$);

\node[draw,circle,below right=0.5cm of pd, inner sep=1pt] (p) {$z^{(l)}_{t+2}$};

\node[minimum size=0, inner sep=0] (g) at (downsample7 -| ppd) {};

\node[right of=g] (upsample) {Upsample};

\node[circle, draw, minimum size=4pt, inner sep=0.1pt] (gu1) at ($(g.center) + (-0.2,-0.6) $) {};
\node[circle, draw, minimum size=4pt, inner sep=0.1pt] (gu2) at ($(g.center) + (0.2,-0.6) $) {};
 
\draw[->] (pd) edge (p) (d) edge (pd);
\draw (pd) edge (ppd)  ;
\draw (p) edge[out=-135, in=90] (ppd);
\draw (ppd) edge (g);

\draw[->] (g) edge[out=-90, in=45] (gu1);
\draw[->] (g) edge[out=-90, in=135] (gu2);

\node[circle, minimum size=4pt, inner sep=0.1pt,draw] (gin) at ($ (pd.north) + (0,1) $) {};

\node[circle, minimum size=4pt, inner sep=0.1pt,left=0.2cm of gu1.south west] () {$g_{t+2}^{(l-1)}$};
\node[circle, minimum size=4pt, inner sep=0.1pt,right=0.2cm of gu2.south east] () {$g_{t+3}^{(l-1)}$};
\node[right=0.2cm of gin] () {$g_{t+2}^{(l)}$};
\node[left=0.2cm of dout] () {$d_{t}^{(l)}$};

\draw[->] (gin) -> (pd.north);   

\end{tikzpicture}}
		\caption{A close-up of the dashed square in \Cref{fig:gen_struct} for the generative model. Here $r$ is the receptive field of the Wavenet model. In this instance $S^{(l)} = 2$. }
		\label{fig:latent_levels}
	\end{subfigure}
	
	\caption{An alternative description of the generative part of the model in \Cref{fig:our_ladder_network} where the temporal dimension is explicitly shown. }
	\label{fig:gen_struct_full}
\end{figure}

In our model we combine temporal convolutional networks and striding as building blocks to instantiate \Cref{fig:deep_seq_ar_v1}. We combine this with a hierarchical VAE where we condition the latent states on the previously observed data in an autoregressive fashion, as visualized in \Cref{fig:gen_struct}. Analogously to the hierarchical VAE, the model can roughly be split into to a bottom-up (from data to more abstract features) and a top-down (from abstract features to predictions) network \citep{Maaloe2019}. The autoregressive features will also be used as features for the approximate posterior, inspired by the work of \citet{Aksan2019} which this work extends. 

The bottom-up network produces a set of features that is divided into hierarchical layers. This network also downsamples the signal in between every layer to capture the multiscale property of the model. To avoid excessive cluttering we use the notation $d^{(l)} = d^{(l)}_{1:S^{(l)}:T}$ to denote the bottom-up features which have been downsampled by a total factor of $S^{(l)}$. The downsample is implemented inside the first WavenetBlock function where the residual connection simply uses strided convolutions and the identity connection use the average. We use superscript $D$ to denote Wavenet with downsampling. 
\begin{subequations}
\begin{align}
	\label{eq:top_down_in}
	d^{(1)}&= \text{Wavenet}(y), \\
	d^{(l)} &= \text{Wavenet}^\text{D}(d^{(l-1)}), \qquad l \neq 1.
\end{align}
\end{subequations}

The features of the top-down network, $h^{(l)}~=~h^{(l)}_{1:S^{(l)}:T}$, follows the hierarchical VAE with an additional conditioning and upsampling. The conditioning is performed by concatenating the output from the previous hierarchical layer, which we denote $g^{(l)} = g^{(l)}_{1:S^{(l)}:T}$, with the delayed features of the bottom-up network, followed by a $1{\times}1$ convolution 
\begin{subequations}
\begin{align}
	\label{eq:top_down_cond}
	h^{(l)} &= \text{Wavenet}_{1{\times}1}([g^{(l)}, \text{Delay}(d^{(l)})]), \qquad l \neq L,\\
	h^{(L)} &= \text{Wavenet}_{1{\times}1}(\text{Delay}(d^{(L)})),
\end{align}
\end{subequations}
where $[\cdot]$ denotes concatenation in feature domain. The Delay function delays $d^{(L)}$ one step at the current downsample level and prepends zeros, i.e.
\begin{align}
	\text{Delay}(d_{1:S:T}) = [0, d_{1:S:T-S}],
\end{align}
here the concatenation is done in time domain. Finally $\text{Wavenet}_{1{\times}1}$ denotes a Wavenet where all the dilated convolutions are replaced with $1{\times}1$ convolutions. 

The prior distribution for the latent variable is given by 
\begin{equation}
	\label{eq:top_down_dist}
	\begin{split}
	p(z^{(l)} \given h^{(l)}) =&\\
	 \mathcal{N} \bigg (z^{(l)}& \bigg \vert 
	\begin{split}
	&\mu = \text{Conv}_{1{\times}1}( h^{(l)}),\\  
	&\Sigma = \text{diag}(\text{softplus}(\text{Conv}_{1{\times}1}(h^{(l)}))^2)
	\end{split} \bigg ),
	\end{split}
\end{equation}
Finally, the output of the top-down layer is calculated as,
\begin{subequations}
\begin{align}
	\label{eq:top_down_out}
	g^{(0)} &= \text{CondWavenet}_{1{\times}1}( h^{(1)},z^{(1)}),\\
	g^{(l-1)} &= \text{CondWavenet}^\text{U}_{1{\times}1}(h^{(l)},z^{(l)}), \,\,\, l \neq 1,
\end{align}
\end{subequations}
thus, the latent variables are used as a conditioning sequence in the Wavenet. The Upsampling here, $\text{CondWavenet}^\text{U}$, is similar to the downsample made as a part of the Wavenet. The identity connection is hear upsampled with nearest neighbor and the residual with strided transpose convolutions. The predictive distribution is chosen as $p(y \given \text{Conv}_{1{\times}1}(g^{(0)}))$. The shape of the predictive distribution is chosen with respect to the problem -- Gaussian for real valued data and Bernoulli for binary data or mixtures thereof.

The approximate posterior is expressed using the previously defined features $d^{(l)}$ and $h^{(l)}$ as
\begin{equation}
	\begin{split}
	q(z^{(l)}) = &\\
	\mathcal{N} \bigg ( &z^{(l)} \bigg \vert
	\begin{split}& \mu = \text{Conv}_{1{\times}1}([h^{(l)},d^{(l)}]),  \\
	& \Sigma = \text{diag}(\text{softplus}( \text{Conv}_{1{\times}1}([h^{(l)},d^{(l)}])^2 )
	\end{split} 
	\bigg ).
	\end{split}
\end{equation}
We do not calculate $q$ as a combination of $p$ and another normal distribution, as advocated by \citet{Sonderby2016} and \citet{Aksan2019}. Instead $q$ is parameterized independently of the common features, similar to \citet{Kingma2016}. 

\Cref{fig:gen_struct,fig:latent_levels} visualize the model in an alternative fashion compared to \Cref{fig:our_ladder_network}. \Cref{fig:gen_struct} resolves the temporal dimension and \Cref{fig:latent_levels} visualizes a hierarchical layer locally in time. The parameters of this model (both the generative model and the approximate posterior) are the parameters for the Wavenets, the Convolutional layers, the Upsample layers and the Downsample layers. 

Although the stochastic TCN (STCN) \citep{Aksan2019} is very related to our model, there are some distinctions. We generalize the STCN with the hierarchical VAE in favor of the Ladder VAE and also extend it by incorporating multiple timescales. \citet{Aksan2019} argue that their model expresses multiple timescales but we disagree since it is missing the downsampling and upsampling steps. Indeed, the deeper features of the STCN gathers information from a larger receptive field. However, these features do only affect the current time step and are not shared between different timesteps.

\section{Experiments}

We conduct experiments on two hierarchically structured sequential datasets, raw waveform speech (Blizzard) and handwritten text (IAM-OnDB). For all experiments we use linear free bits (see \Cref{sec:hyperparameters}) and the Adam optimizer with an exponentially decaying learning rate of $5\cdot10^{-4}$. For free bits we used a scheme to smoothly decay the threshold of free bits exponentially over the course of training at a rate of halving the threshold every $300\thinspace000$ iterations. We train for $100$ epochs and do not see any tendencies of overfitting on the training data. Finally, each bottom-up Wavenet consists of four Wavenet blocks and each top-down constsits of two Wavenet blocks. The number of Wavenet blocks are thus the same in the bottom-up and the top-down networks. For exact hyperparameter settings and additional implementation details we refer to the appendix and code\footnote{\url{https://github.com/carl-andersson/MS-STCN}}.

{\bf Blizzard} is a raw waveform speech dataset that originates from the 2013 Blizzard challenge \citep{Karaiskos2013}. The raw signal is split into $0.5$ second ($8\thinspace000$ samples) long sequences. We experiment by dividing these sequences into frames of $200$ samples down to $2$ to really enable the long memory requirements and hierarchical decomposition of the data. Previous research on this dataset used frames of $200$ samples. Additionally, we remove some artifacts from the mp3 decoding that appear before and after every sequence. The remaining preprocessing is performed according to \citet{Aksan2019}. 

The smaller framesize increases the computational complexity of the model compared to that of the STCN if we were to use the same number of filters. We compensate for this by decreasing the number of filters in our model for layers with small $l$ compared to those of the STCN. This way the computational complexity is approximately the same for the all of proposed models as for the (moderate) STCN, while at the same time using significantly fewer parameters. In the case of frame width of 200 we keep all the filters the same size which to be comparable with STCN. 

The ELBO evaluated on test data for the Blizzard dataset is presented in \Cref{tab:results_speech_text}. Our model without striding is essentially a reimplementation of STCN using the model setup described in this paper. As a control experiment we also report the result with our model without the stochastic latent variables (i.e. only deterministic autoregressive). However, this control showed instabilities and eventually diverged so we report the best fit we got before it diverged.  This divergence might be an indication that the latent variables are somehow regularizing the model. In \Cref{tab:results_speech_text_kl} we also report the amount of information that is used in the KL-term for the different layers. Results show that KL units on all levels are being used. 

{\bf IAM-OnDB} \citep{Liwicki2005_iamondb} is a handwritten text dataset where every sample consists of the pen position ($\mathbb{R}^2$) and a binary variable indicating whether the pen was lifted after this position or not. We use the same preprocessing as \citet{Chung2015} and \citet{Aksan2019}. However, we noted that the discretization of the pen position in some cases led to models that greatly overestimated the likelihood fit to the data. To cope with these discretization artifacts  we added artificial uniform measurement noise to the pen position. 

Samples from the model trained on IAM-OmDB (\Cref{fig:samples}) shows complex sequences were it is even possible to recognize a few words. However, the added noise means that the likelihood estimates of the model are not easily comparable with previous results on this dataset. 

\begin{table}[ht]
	\sisetup{table-format = <5, table-number-alignment = center, mode=text}
	\begin{center}     
		\captionsetup{width=.9\linewidth}
		\caption{Average ELBO per sequence comparison for speech with varying frame width (FW) and with and without down-/upsampling (multiscale). }
		\label{tab:results_speech_text}
		{ \small
			\begin{tabular}{l S[table-format = <5] S[table-format = <4] S[table-format = <4]}
				\toprule
				Model & \multicolumn{1}{c}{Blizzard} \\
				\midrule
				RNN (GMM)  \citep{Chung2015} & 7413 \\
				Wavenet (GMM) \cite{Aksan2019} & 8190 \\
				VRNN (GMM) \citep{Chung2015} & \approx 9392 \\
				SRNN \citep{Fraccaro2016} & \geq 11991 \\
				Z-Forcing \citep{Goyal2017} & \geq 15430 \\
				STCN (Moderate) \citep{Aksan2019} & \geq 16288 \\
				STCN (Large) \citep{Aksan2019} &  \geq 17670 \\
				\midrule
				Our model, FW=200 & \geq 13645 \\ 
				Our model, FW=200 w/o multiscale  & \geq 15746 \\
				Our model, FW=25 & \geq 17060 \\ 
				Our model, FW=25 w/o multiscale   & \geq 16562  \\
				Our model, FW=2 &  \B \geq 18809 \\ 
				Our model, FW=2 w/o multiscale & \geq 18372 \\
				Our model, FW=2 w/o latent variables & 18523 \\
				\bottomrule
			\end{tabular}
		}
	\end{center}
\end{table}

\vspace{-0.5em}
\section{Conclusion and future work}
\vspace{-0.5em}

In this paper we argue that a deep learning model is not that different from a structured autoregressive model. However, the philosophy behind the model and the problems it is applied to, are significantly different from a typical system identification model. The model we proposed has shown to model hierarchically structured data very well while at the same time being more parameter efficient compared to existing models. We hypothesize that the proposed model exploit the hierarchical nature of the sequential data similar to how the CNN exploits the hierarchical structure in image data. 

The proposed model is in principle not limited to one-dimensional temporal data. Similar to PixelVAE \citep{Gulrajani2016}, our model can be applied to images or even video. Investigating whether the promising results of this model transfers into other regimes of hierarchically structured data is an interesting avenue for future work. Another area for further experiments is to apply the model to other hierarchically structured data discussed in the introduction, one such example is pose and orientation estimation from inertial sensors \citep{Manon2017}.

\begin{table}[ht]
	\sisetup{table-format = 3, table-number-alignment = center, mode=text, round-precision=0, round-mode = places}
	\begin{center}     
		\captionsetup{width=.9\linewidth}
		\caption{Average KL-divergence per layer on the Blizzard dataset where KL1 corresponds to the layer closest from the data. }
		\label{tab:results_speech_text_kl}
		{ \small
			\begin{tabular}{l S[table-format = 3] S[table-format = 3] S[table-format = 3] S[table-format = 3] S[table-format = 3]}
				\toprule
				Model & \multicolumn{1}{c}{KL1} & \multicolumn{1}{c}{KL2} & \multicolumn{1}{c}{KL3} & \multicolumn{1}{c}{KL4} & \multicolumn{1}{c}{KL5}\\
				\midrule				
				FW=200 & 0.5849201800005052& 7.99086228375918& 4.451699510602553& 3.931846768133616& 74.69217626576035 \\ 
				FW=200 w/o multiscale & 48.63927638095835 & 33.65645178274352 & 47.96300490885383 & 146.00738400771718 & 142.97979292863562 \\
				FW=25 & 235.4 & 45.28940935 & 44.33512293 & 66.62584468 & 101.70034407 \\
				FW=25 w/o multiscale  &  320.93469148379603 & 126.59908707128865 & 488.13101236620463 & 542.9513418013241 & 180.42520826180387 \\ 
				FW=2 &  629.1813384740739 & 48.9450172994962 & 45.18734085857684 & 35.25978321085142 & 57.97964441838387 \\ 
				FW=2 w/o multiscale & 203.5077593894518 & 210.2916540831487 & 158.27784056474766 & 209.4174622465976 & 50.33677824322617 \\
				
				\bottomrule
			\end{tabular}
		}
	\end{center}
\end{table}

\section*{Acknowledgment}
\vspace{-0.5em}

We thank Joakim Jaldén's group at KTH, Royal Institute of Technology, Stockholm, for lending computational power.

\bibliography{references}
	
\newpage

\appendix

\section{Hyperparameters and Implementation details}
\label{sec:hyperparameters}
Code is implemented in Pytorch \citep{Paszke2017}. All experiments are done with a Titan Xp except the final experiment on the Blizzard dataset which was done with a Tesla V100. 

\subsection{Hyperparmeters}

This section specifies the hyperparameters for the different models used. All models consists  of 5 hierarchical layers. However the number of filters are chosen to match that the computational load/number of parameters in STCN.

{\bf Blizzard:} The Wavenets in the bottom-up network consists of 5 Wavenet blocks  with kernel size 2 and the Wavenets in the top-down network consts of 2 Wavenet Blocks each. The dilation rates for the bottom-up networks where chosen as $[1,2,4,8,16]$ when not using downsampling and $[]1,1,1,1,1]$ otherwise. An exception to this is for frame widths 25 and 2 with downsampling as the topmost bottom-up layer then uses $[1,2,4,8,16]$

{\bf Frame width 200 and multiscale} 256 filters in all convolutions with 64 latent variables for each VAE layer and timestep. The downsampling is $[1,2,2,2,1]$ for the VAE layers respectively and the predictive distribution is chosen as a mixture of 10 Gaussians.  

{\bf Frame width 200 and NO multiscale} 256 filters in all convolutions with 64 latent variables for each VAE layer and timestep. The downsampling rate is $[1,1,1,1,1]$ for the VAE layers respectively and the predictive distribution is chosen as a mixture of 10 Gaussians.

{\bf Frame width 25 and multiscale} $[96, 128, 192, 256, 256]$ filters the convolutions in respective VAE layer and $[24, 32, 48, 64, 64]$ latent variables for each VAE layer and timestep. The downsampling rate is $[1,2,2,2,1]$ for the VAE layers respectively and the predictive distribution is chosen as a mixture of 10 Gaussians.

{\bf Frame width 25 and NO multiscale} 96 filters for all the convolutions in respective VAE layer and 24 latent variables for each VAE layer and timestep. The downsampling rate is $[1,1,1,1,1]$ for the VAE layers respectively and the predictive distribution is chosen as a mixture of 10 Gaussians.

{\bf Frame width 2 and multiscale} $[26, 58, 128, 256, 256]$ filters the convolutions in respective VAE layer and $[1, 5, 25, 64, 64]$ latent variables for each VAE layer and timestep. The downsampling rate is $[1,5,5,4,1]$ for the VAE layers respectively and the predictive distribution is chosen as a mixture of 2 Gaussians.

{\bf Frame width 2 and NO multiscale} 26 filters for all the convolutions in respective VAE layer and 1 latent variables for each VAE layer and timestep. The downsampling rate is $[1,1,1,1,1]$ for the VAE layers respectively and the predictive distribution is chosen as a mixture of 2 Gaussians.

The number of filter for each examples is chosen to keep the computational complexity equal to that of the moderate STCN.

{\bf IAM-OnDB} The Wavenets in the bottom-up network consists of 4 Wavenet blocks  with kernel size 2 and the Wavenets in the top-down network consts of 2 Wavenet Blocks each. The dilation rates for the bottom-up networks where chosen as $[1,1,1,1,1]$ except for the topmost layer where we used $[1,2,4,8]$. 64 filters in all convolutions with $[8,8,8,4,2]$ latent variables for VAE layer respectively and timestep. The downsampling is $[1,2,2,2,1]$ for the VAE layers respectively and the predictive distribution is chosen as a mixture of 10 Gaussians and a Bernoulli for the pen up event.  

\subsection{Predictive distributions}

The final distribution is parameterized as $p(y \given z^{(1)}) = p(y \given \text{Conv}_{1{\times}1}(g^{(0)}))$. Let us denote the features $\text{Conv}_{1{\times}1}(g^{(0)})$ with $\eta$ for some clarity, we also denote the dimension of $y$ with $p$. 

\paragraph{Gaussian} We parameterize the final prediction the same way as the latent variables. The features $\eta$ is split up into mean features, $\eta_\mu$ and standard deviation features, $\eta_\sigma$. In addition to what is mentioned in the paper we clamp the standard deviation of the Gaussian distribution between $10^{-3}$ and $5$, both for the final prediction and the latent variables. The distribution is thus given of 
\begin{align}
	p(y \given z^{(1)}) = \mathcal{N}(y \given \eta_\mu , \text{softmax}(\text{Clamp}(\eta_\sigma, 10^{-3}, 5.0))^2),
\end{align}
and the dimension of $\eta$ must thus be $2p$. 

\paragraph{Bernoulli} For the Bernoulli distribution we simply parameterize with the logits which is equal to $\eta$. The dimension for $\eta$ is thus $p$. 

\paragraph{Mixtures} For mixture distributions with $n_c$ components we split $\eta$ into $n_c$  weight parameters and the rest to parameterize the $n_c$ base distributions. The weight parameters are interpreted as the logit weights. The dimension for $\eta$ in this case is $2 p n_c+ n_c$ for the Gaussian mixture model and $p n_c + n_c$ for the Bernoulli mixture model.

\subsection{Posterior collapse}
Without any additional tricks, the hierarchical VAE suffers from \emph{posterior collapse}, a state during the optimization where the model gets stuck in a local minimum and the KL units do not transfer any information  \citep{Sonderby2016, Kingma2016, Goyal2017, Bowman2016}. This state can be circumvented through a number of different strategies, e.g. \emph{KL-annealing} \citep{Sonderby2016, Bowman2016}, \emph{free bits} \citep{Kingma2016} or auxiliary loss functions as in Z-forcing \citep{Goyal2017}. 

To avoid posterior collapse we use an alternative to free bits \citep{Kingma2016}. The ordinary free bits sets a minimum value of how low the KL units can be minimized by altering the ELBO objective \Cref{eq:elbo} to
\begin{equation}
\mathcal{L}_\text{FB} = \expect_q \log p(y \given z) - \sum_i \text{maximum} \left ( D_\text{KL} \left ( q(z_i) || p(z_i) \right ), \lambda \right ),
\end{equation}  
where $\lambda$ denotes the free bits threshold and $i$ iterates through all latent variables, $z$. To smoothen the transition from where the threshold is active to where it is not, we propose a linear smoothing of the free bits objective. Hence, we define linear free bits
\begin{equation}
\begin{split}
\mathcal{L}&_\text{LFB} = \expect_q \log p(y \given z) - \\ &\sum_i  \Psi \left ( \text{min} \left ( \frac{D_\text{KL} \left ( q(z_i) || p(z_i) \right ) }{\lambda}  , 1  \right )   \right )  D_\text{KL} \left ( q(z_i) || p(z_i) \right ),
\end{split}
\end{equation}  
where $\Psi(\cdot)$ is a stop gradient function that ensures that no gradient information is passed through the function in the backward pass while acting as the identity function in the forward pass.

\end{document}